\documentclass[10pt,twocolumn,letterpaper]{article}

\usepackage{3dv}
\usepackage{times}
\usepackage{epsfig}
\usepackage{graphicx}
\usepackage{amsmath}
\usepackage{amssymb}
\usepackage{graphicx}
\usepackage{multirow}
\usepackage{booktabs}
\usepackage{mathtools}
\usepackage{bm}
\usepackage{bbm}
\usepackage{pdfpages}
\usepackage{multido}

\usepackage[pagebackref=true,breaklinks=true,colorlinks,bookmarks=false]{hyperref}

\usepackage[capitalise]{cleveref}
\Crefname{table}{Tab.}{Figs.}
\Crefname{section}{Sec.}{Figs.}
\Crefname{section}{Eq.}{Figs.}

\providecommand{\eg}{\textit{e.g.}\@\xspace}
\providecommand{\ie}{\textit{i.e.}\@\xspace}

\newcommand{\mtosc}{ModelNet$\rightarrow$ScanNet}
\newcommand{\stom}{ShapeNet$\rightarrow$ModelNet}
\newcommand{\stosc}{ShapeNet$\rightarrow$ScanNet}
\newcommand{\sctom}{ScanNet$\rightarrow$ModelNet}
\newcommand{\sctos}{ScanNet$\rightarrow$ShapeNet}
\newcommand{\algoname}{RefRec}
\newcommand{\pl}{pseudo-labels}
\newcommand{\pt}{pre-training}
\newcommand{\Pt}{Pre-training}
\newcommand{\st}{self-training}

\threedvfinalcopy 

\def\threedvPaperID{****} 

\ifthreedvfinal\fi
\begin{document}

\title{\algoname{}: Pseudo-labels Refinement via Shape Reconstruction \\ for Unsupervised 3D Domain Adaptation}

\author{Adriano Cardace\qquad Riccardo Spezialetti\qquad Pierluigi Zama Ramirez\qquad \\
Samuele Salti\qquad Luigi Di Stefano\\
Department of Computer Science and Engineering (DISI)\\
University of Bologna, Italy\\
{\tt\small \{adriano.cardace2, riccardo.spezialetti, pierluigi.zama\}@unibo.it}
}

\maketitle
\thispagestyle{empty}

\begin{abstract}
Unsupervised Domain Adaptation (UDA) for point cloud classification is an emerging research problem with relevant practical motivations. Reliance on multi-task learning to align features across domains has been the standard way to tackle it.
In this paper, we take a different path and propose \algoname{}, the first approach to investigate \pl{} and \st{} in UDA for point clouds. We present two main innovations to make \st{} effective on 3D data: i) refinement of noisy \pl{} by matching shape descriptors that are learned by the unsupervised task of shape reconstruction on both domains; ii) a novel self-training protocol that learns domain-specific decision boundaries and reduces the negative impact of mislabelled target samples and in-domain intra-class variability.
\algoname{} sets the new state of the art in both standard benchmarks used to test UDA for point cloud classification, showcasing the effectiveness of \st{} for this important problem.
\end{abstract}

\section{Introduction}
\label{sec:intro}

Properly reasoning on 3D geometric data such as point clouds or meshes is crucial for many 3D computer vision tasks, which are key to enable emerging applications like autonomous driving, robotic perception and augmented reality. In particular, assigning the right semantic category to a set of points that represent the surface of an object is a required skill for an intelligent system in order to understand the 3D scene around it. Such problem, referred to as \textit{shape classification}, was initially addressed by \textit{hand-crafted} features \cite{kazhdan2003rotation,arras2007using,salti2010use}, while, with the advances in deep learning, recent proposals learn features directly from 3D point coordinates by means of deep neural networks \cite{qi2017pointnet,qi2017pointnet++, liu2020closer,yan2020pointasnl,xu2020grid,liu2019relation,hua2018pointwise,wang2019dynamic,thomas2019kpconv,li2018so}. Although data-driven approaches can achieve impressive results, they require massive amounts of labeled data to be trained, which are cumbersome and time-consuming to collect. Typically, 3D deep learning methods use synthetic datasets of CAD models, \eg ModelNet40 \cite{wu20153d} or ShapeNet \cite{chang2015shapenet}, to harvest a large number of 3D examples. While synthetic datasets enable 3D deep learning, they create a conundrum. On the one hand, shape classifiers trained on ModelNet are very effective on synthetic data, as witnessed by performance saturation on standard benchmarks \cite{uy2019revisiting,goyal2021revisiting}. On the other hand, though, they are not able to transfer their performance to real-world scenarios \cite{uy2019revisiting}, where point cloud data are usually captured by RGB-D or LiDAR sensors \cite{dai2017scannet,hua2016scenenn}. This limitation severely restricts the deployment of  3D deep learning methods in real-world applications.

\begin{figure}[t]
	\centering
	\includegraphics[trim={8cm 6cm 8cm 5cm}, clip, scale=0.4]{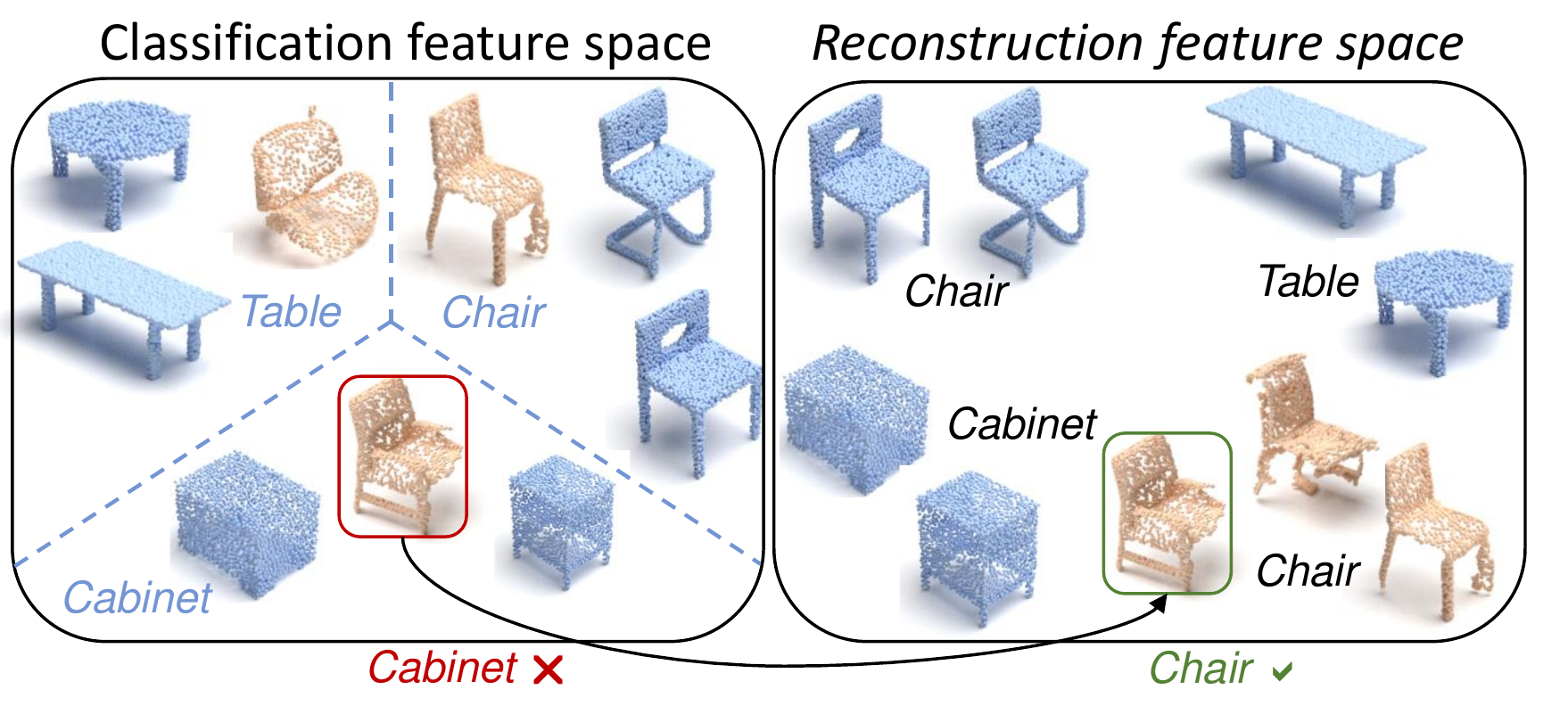}
	\caption{Feature space of a classifier (left) and a reconstruction auto-encoder (right). A classifier trained on source shapes only (blue) may not be effective on target shapes (orange) and assign wrong \pl{}. An auto-encoder, instead, tends to cluster similar shapes together in the learned embedding, such that  its features can be used as global shape descriptors to correct wrong \pl{}.}
	\label{fig:teaser}
\end{figure}

Unsupervised Domain adaptation (UDA) aims at bridging this domain shift \cite{shift} by learning to transfer the knowledge gained on a labeled dataset, \ie{} \emph{source domain}, to an unlabeled target dataset, \ie{} \emph{target domain}. 

UDA has its roots in 2D computer vision, where a multitude of methods have been proposed \cite{kouw2019review}. Among them, the most widespread approach pertains globally aligning the feature distributions between the source and target domain. 
This is the paradigm also leveraged by methods tackling UDA for 3D data, either explicitly, by designing losses and models to align features \cite{qin2019pointdan}, or implicitly, by solving a self-supervised task on both the source and target domains with a shared encoder \cite{alliegro2021joint, achituve2021self}.
We argue that, when moving from a synthetic CAD dataset to a real world one, feature alignment can only lead to sub-optimal solutions due to the large differences between the two domains. Indeed, acquiring objects in cluttered scenes results oftentimes in partial scans with missing parts due to \textit{occlusions}. Moreover, registration errors arising when fusing multiple 2.5D scans \cite{dai2017bundlefusion} to obtain a full 3D shape, alongside noise from the sensor, result in less clean and geometrically coherent shapes than CAD models (see Fig. \ref{fig:intra_variation}). Therefore, the shape classifier may need to ground its decision on new cues, different from those learned on the source domain, where the clean full shape is available, to correctly classify target samples.

To let the classifier learn such new cues, in this paper we propose \algoname{} (\underline{Ref}inement via \underline{Rec}onstruction), a novel framework to perform unsupervised domain adaptation for point cloud classification. Key to our approach is reliance on \textit{\pl{}} \cite{lee2013pseudo}, \ie{} predictions on the target domain obtained by running a model trained on the source domain, which are then used as a noisy supervision to train a classifier on the target domain and let it learn the domain-specific cues. This process is usually referred to as \textit{self-training} \cite{lee2013pseudo, zou2018unsupervised}.
However, the \pl{} obtained from a model trained on the source domain may be wrong due to the domain shift (as shown in Fig. \ref{fig:teaser}-left) and a target domain classifier naively trained on them would underperform. Therefore, our key contribution concern effective approaches to \emph{refine} \pl{}. We propose both an offline  and an online refinement, \ie before training and while training on \pl{}.  Both refinements are based on the idea that similar shapes should share similar labels. To find similar shapes, we match \emph{global shape descriptors}, \ie the embedding
computed by an encoder given the input shape. Here we make another key observation, illustrated in Fig. \ref{fig:teaser}: the space of features learned by a classifier is organized to create linear boundaries among different classes, but it is not guaranteed -nor meant- to posses a structure where similar shapes lay close one to another, especially for target samples which are not seen at training time. Hence, such features are not particularly effective if used as global shape descriptors. In contrast, teaching a \textit{point cloud auto-encoder} to reconstruct 3D shapes is an effective technique to obtain a compact and distinctive representation of the input geometric structures, as proven by recent proposals for local and global shape description \cite{yang2018foldingnet,groueix2018,deng2018ppf,spezialetti2019learning}, which, in our setting, can be trained also on the target domain since it is learned in an unsupervised way. By leveraging on such properties of the reconstruction latent space, in the offline step we focus on reassigning the \pl{} of target samples where the source domain classifier exhibits low confidence, while in the online step we compute \textit{prototypes} \cite{Pinheiro2018UnsupervisedDA}, \ie the mean global descriptors on the target domain for each class, and we weight target \pl{} according to the similarity of the input shape to its prototype. Peculiarly, by using reconstruction embeddings trained also on the target domain to compute prototypes, we avoid the domain shift incurred when using the classifier trained on the source domain as done by previous 2D methods \cite{kang2019contrastive}. In the online refinement step, we also leverage the standard training protocol of 2D UDA methods based on mean teacher \cite{ema} to improve the quality of \pl{} as training progresses.

We can summarize our contributions as follows:
\begin{enumerate}
    \item we investigate on \st{} to solve UDA for point clouds, an approach that sharply differs from existing proposals in literature based on multi-task learning. To the best of our knowledge, this work is the first to study this alternative path;

    \item we show how global descriptors learned for shape reconstruction can be effectively used both offline and online to refine \pl{} in UDA for point cloud classification;
    
    \item we show how effective techniques for 2D UDA, like domain-specific classifiers and mean teacher supervision, can be successfully used on 3D data; 
    
   \item we achieve new state of the art performance on the standard benchmarks used to assess progress in UDA for point cloud classification.
\end{enumerate}

\begin{figure*}[t]
    \centering
    \includegraphics[trim={0cm 2.5cm 0cm 3cm}, clip, width=0.95\linewidth]{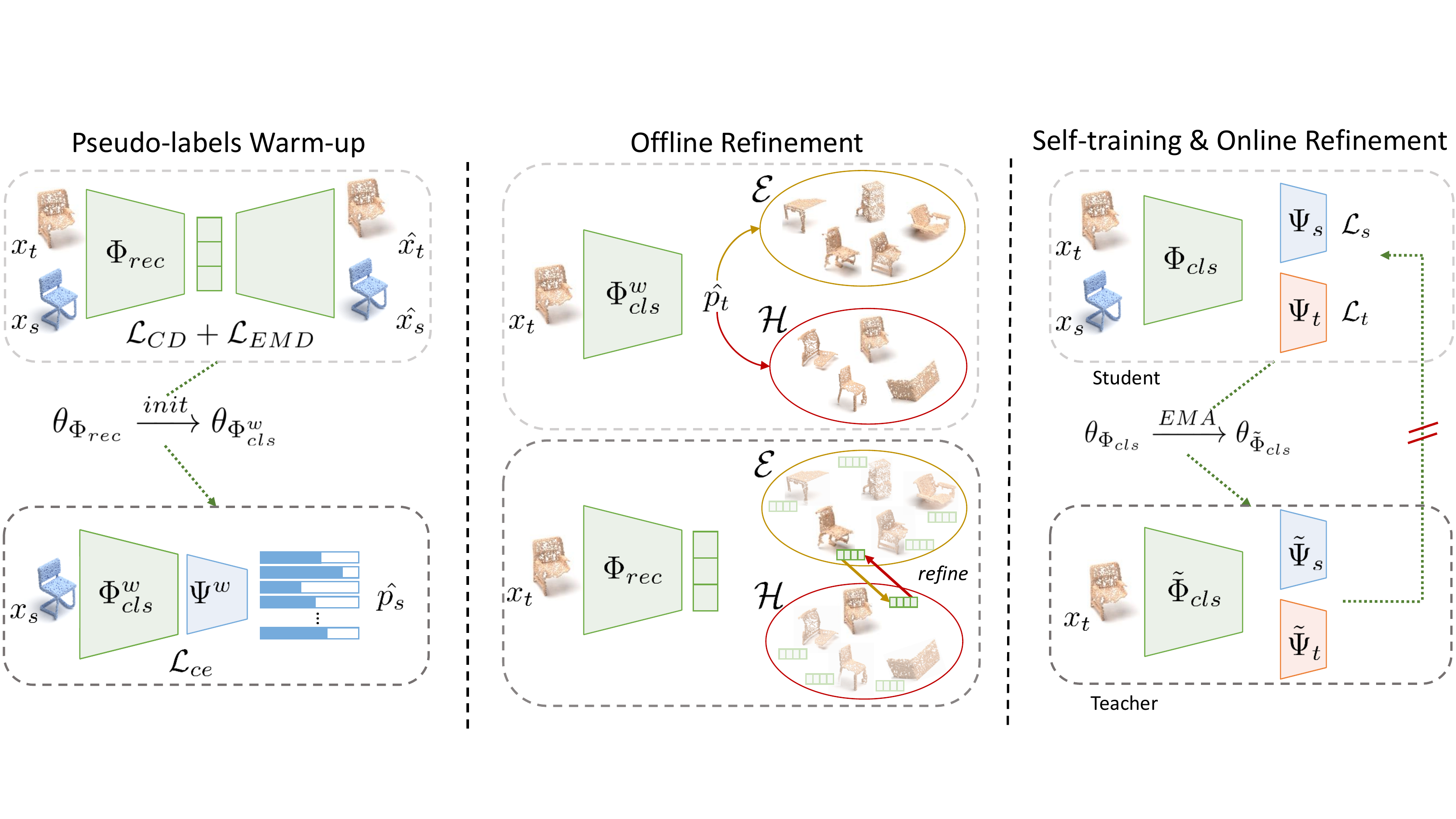}
    \caption{\algoname{} comprises three steps. First, in the \textit{pseudo-labels warm up} step, we train a reconstruction network $\Phi_{rec}$ on both source and target domains. The weights of the encoder are used to initialize the backbone $\Phi^{w}_{cls}$ of a classifier, that is then trained on the source domain.
    In the \textit{refinement} step, we use the classifier to split target samples in the easy ($\mathcal{E}$) and hard ($\mathcal{H}$) sets according to their confidence and refine them by performing nearest neighbor queries in the auto-encoder feature space.
    Finally, in the \textit{self-training} step, we train a target-specific classifier $\Psi_t$ by refined \pl{} and online \pl{} obtained with the mean teacher architecture \cite{ema}.
    }
    \label{fig:pipeline}
\end{figure*}

\section{Related Work}
\label{sec:related}

\subsection{Unsupervised Domain Adaptation (UDA)}
\label{sec:uda}
Unsupervised Domain Adaptation aims at reducing the need of large amounts of annotated data. The key idea is to learn distinctive and powerful features in the source domain and exploit such representation in the target domain. A remarkable amount of work has been conducted for image classification \cite{ganin2015, Bousmalis2016, Long2017, NIPS2016_ac627ab1, Sun2016DeepCC, Tzeng2017, Tzeng2014DeepDC, Li2017RevisitingBN}, semantic segmentation \cite{bdl,  adaptsegnet, maxsquare, pmlr-v80-hoffman18a, DCAN,  ltir} and object detection \cite{Chen2018,  Wang_2019_CVPR, Wang_2019_CVPR, Wang_2019, Saito2019}.
The most common approach to tackle UDA is to minimize the discrepancy between domains to obtain domain-invariant features, so that the same classifier can be deployed in both domains. Alignment in feature space can be achieved by forcing features from both domain to have similar statistics, as done in \cite{Long2017, NIPS2016_ac627ab1}. Another interesting line of research, instead, casts domain alignment as a min-max problem, exploiting adversarial training to attain such alignment. For example, Tzeng et al. \cite{Tzeng2017} introduced a domain confusion loss to obtain  features indistinguishable across domains.
Ganin et al. \cite{ganin2015} try to learn domain invariant representations in an adversarial way by back-propagating the reverse gradients of the domain classifier.
All these methods are meant to work with images and do not explore any variations or extensions to  exploit 3D data. In this work, instead, we turn our attention on 3D data and focus on their peculiar properties to design an effective domain adaptation methodology.

\subsection{Unsupervised 3D Domain Adaptation}
\label{subsec:da_pcd}

Only few papers discuss Unsupervised Domain Adaptation for point cloud classification. Among these, PointDAN \cite{qin2019pointdan} is a seminal work that proposes to adapt a classical 2D domain adaptation approach to the 3D world. Specifically, they focus on the alignment of both local and global features, building their framework upon the popular Maximum Classifier Discrepancy (MCD) \cite{Saito2018MaximumCD} for global feature alignment.
Differently, \cite{alliegro2021joint,achituve2021self} leverage on Self-supervised learning (SSL) to learn simultaneously distinctive features for both the source and target domains.
Similarly, \cite{alliegro2021joint} also relays on SSL, and introduces a novel pretext task to learn strong features for the target domain as well.
In this work, we take one step further and present an unsupervised domain adaptation method based on \pl{} that exploits a shape reconstruction task to refine them.

\subsection{Deep Learning for Point Clouds Reconstruction}
\label{subsec:ssl_pcd}
With the recent advances in deep learning several methods for point cloud reconstruction have been suggested.  A seminal work in this area \cite{Achlioptas2018LearningRA} proposes 
a new auto-encoder architecture for point clouds using the permutation invariant operator introduced in \cite{qi2017pointnet}. AtlasNet \cite{groueix2018} and FoldingNet \cite{yang2018foldingnet} propose a \textit{plane-folding} decoder to learn to deform points sampled from a plane in order to reconstruct the input surface. TearingNet \cite{pang2021tearingnet} takes inspiration from \cite{yang2018foldingnet,groueix2018} and present a \textit{tearing} module to cut regular 2D patch with holes, or into several parts, to preserve the point cloud topology. In \cite{chen2019unpaired}, the authors reconstruct the point clouds by training a generative adversarial network (GAN) on the latent space of unpaired clean synthetic and occluded real point clouds. We take inspiration from the finding of \cite{Achlioptas2018LearningRA} and leverage the expressive power of point cloud auto-encoders to pre-train our shape classifier and learn a global shape descriptor deployed for \pl{} refinement.

\section{Method}
\label{sec:method}

In this work, we address UDA for point clouds classification. Hence, we assume availability of a labeled source domain $\mathcal{S} = {\{(\textbf{x}_s^i \in \mathcal{X}_s, y_s^i \in \mathcal{Y}_s)\}_{i=1}^{n_s}}$, and a target domain $\mathcal{T} = \{\textbf{x}_t^j \in \mathcal{X}_t\}_{j=1}^{n_t}$, whose labels $\{y_t^j \in \mathcal{Y}_t\}_{j=1}^{n_t}$ are, however, not available. As in standard UDA settings \cite{Wang_2018}, we assume to have the same one-hot encoded label space $\mathcal{Y}_s = \mathcal{Y}_t =\mathcal{Y} = \{0,1\}^k$ and the same input space $\mathcal{X}_s = \mathcal{X}_t$ (\ie{} point clouds with $\{x,y,z\}$ coordinates) but with different distributions $P_s(\textbf{x}) \neq P_t(\textbf{x})$, \eg due to source data being synthetic while target data being real or due to the use of different sensors. 
The final classifier for a point cloud $\textbf{x}$ can be obtained as a composite function $\Omega=\Phi \circ \Psi$, with $\Phi: \mathcal{X} \rightarrow \mathbb{R}^{d}$ representing the feature extractor and $\Psi: \mathbb{R}^{d} \rightarrow [0,1]^k$ the classification head, which outputs softmax scores $\hat{\mathbf{p}} \in [0,1]^k$. When one-hot labels are needed, we further process softmax scores with $\Lambda: \mathbb{R}^k \rightarrow \mathcal{Y}$ to obtain the label corresponding to the largest softmax score, with such value 

 providing also the confidence associated with the label prediction. Although the largest softmax is a rather naive confidence measure, we found it to work satisfactorily in our experiments. Our overall goal is to learn a strong classifier $\Omega_t$ for the target domain even though annotations are not available therein.

An overview of our method is depicted in \cref{fig:pipeline}. It encompasses three major steps: warm-up, \pl{} refinement, and self-training. The purpose of the first step, described in \cref{sec:warmup}, is to train a  model effective  on target data by using labelled source data and unlabelled target data. Once trained, this model provides the initial \pl{}. 
These are refined offline in the second step, described in \cref{sec:ref}, in order to partially reduce the errors in \pl{} due to the limited generalization of the source classifier to the target domain.
Finally, in the last step, detailed in \cref{self_training}, we introduce an effective way of exploiting \pl{} during \st{} by combining a domain-specific classifier with an online \pl{} weighting strategy that exploits prototypes computed in the target domain.

\subsection{Pseudo-labels Warm-up}
\label{sec:warmup}
The first step of our pipeline seeks to produce good initial \pl{}, which, after refinement, can be used to train the final classifier on the source domain and the target domain augmented with \pl{}. The warm-up step is very important, as the effectiveness of \st{} is directly related to the quality of \pl{}.
To obtain good initial \pl{}, we focus on pretraining and data augmentation, with the aim of reducing  overfitting on source data.
\Pt{} is largely adopted also in UDA for image classification, where ImageNet \pt{} is a standard procedure \cite{kang2019contrastive, Pei2018MultiAdversarialDA, Long2017} that learns powerful features able to generalize to multiple domains and alleviates the risk of overfitting when training solely on data coming from the source distribution.
Differently from UDA in the 2D world, however, here we focus on \emph{unsupervised} pre-training. This is particularly attractive for the UDA context, where no supervision is available for the target domain. In fact, inspired by recent advances on representation learning for 3D point clouds, which have demonstrated the effectiveness of unsupervised techniques for learning discriminative features \cite{yang2018foldingnet,groueix2018,deng2018ppf,spezialetti2019learning}, we propose to use point cloud reconstruction as unsupervised \pt{} for our backbone. The key advantage of such \pt{} is the possibility to capture discriminative features also for the target domain  since unsupervised \pt{} can be conducted on both domains simultaneously. Moreover, it learns a feature extractor $\Phi_{rec}$  which can be deployed also to refine labels effectively, as we do in the following steps of our pipeline.

We follow the same strategy proposed in \cite{Achlioptas2018LearningRA}, and use a standard PointNet \cite{qi2017pointnet} backbone as $\Phi_{rec}$ to produce a global $d$-dimensional descriptor of the input point cloud. This latent representation is then passed to a simple decoder made out  of 3 fully connected layers that tries to reconstruct the original shape. During training we minimize both the Chamfer Discrepancy (CD) $\mathcal{L}_{CD}$ and Earth Mover’s distance (EMD) $\mathcal{L}_{EMD}$ \cite{rubner2000earth} as loss functions \cite{Achlioptas2018LearningRA}. Additionally, as mentioned above, data augmentation is a key ingredient to improve generalization, especially for the synthetic-to-real adaptation case, since 3D real scans always exhibits occlusions and non-uniform  point density. For this reason, when performing synthetic-to-real adaptation, we apply a  data augmentation procedure similar to that  proposed in \cite{spezialetti2020learning} in order to simulate occlusions. 

To conclude warm-up, we train a new classifier $\Omega^{w}=\Phi^{w}_{cls} \circ \Psi^{w}$ on the source dataset with a classical cross-entropy loss. Importantly, $\Phi^{w}_{cls}$ and $\Phi_{rec}$ have the same architecture and the weights $\theta$ of the backbone ${\Phi^{w}_{cls}}$ are initialized with those learned for $\Phi_{rec}$. We then use $\Omega^{w}$ to obtain   \pl{} $\{\hat{y}^j_t = \Lambda ( \Omega^{w}(\mathbf{x}^j_t) \}_{j=1}^{n_t} )$ alongside their confidence scores.

We may, in principle, exploit these   \pl{} to perform self-training in the target domain.  However, even if we rely on unsupervised \pt{} and data augmentation to boost performance on the target domain, they  are still  noisy. Indeed, due to the domain gap, only a small portion of the  \pl{} can be considered reliable, while the majority of the samples are assigned wrong labels that could lead to poor performance when applying \st{}. Hence, in the next step, we refine the initial \pl{} obtained in the warm-up step by leveraging on $\Phi^{w}_{rec}$.

\subsection{Pseudo-labels Refinement}
\label{sec:ref}
To refine \pl{}, we exploit the confidence computed by the classifier $\Omega^{w}$ and split \pl{} in the two disjoint sets of highly confident predictions, denoted as $\mathcal{E}$ (\ie{} \textit{easy split}), and uncertain ones, denoted as $\mathcal{H}$ (\ie{} \textit{hard split}).
We first build $\mathcal{E}$ by selecting the \textit{g=10}\% most confident predictions on the target samples for each class. We perform this operation class-wisely to obtain a sufficient number of examples for each class and to reproduce the class frequencies in $\mathcal{E}$.
$\mathcal{H}$ is composed by all the remaining target samples.

One of the key idea behind \algoname{} is to utilize the embedding of the reconstruction backbone $\Phi_{rec}$, instead of $\Phi^{w}_{cls}$, to improve the labels of the samples in $\mathcal{H}$. We conjecture that since $\Phi^{w}_{cls}$ has been trained only on the source domain, its embeddings are not discriminative for the target domain, and more importantly there are no guarantees that objects belonging to the same class, yet coming from different domains, would lay close in the feature space. Hence, we assign new \pl{} to target samples according to similarities in the feature space of $\Phi_{rec}$.
\begin{figure}[t]
    \centering
    \includegraphics[
    trim={6cm 5cm 6cm 2cm}, clip,
    width=0.7\linewidth]{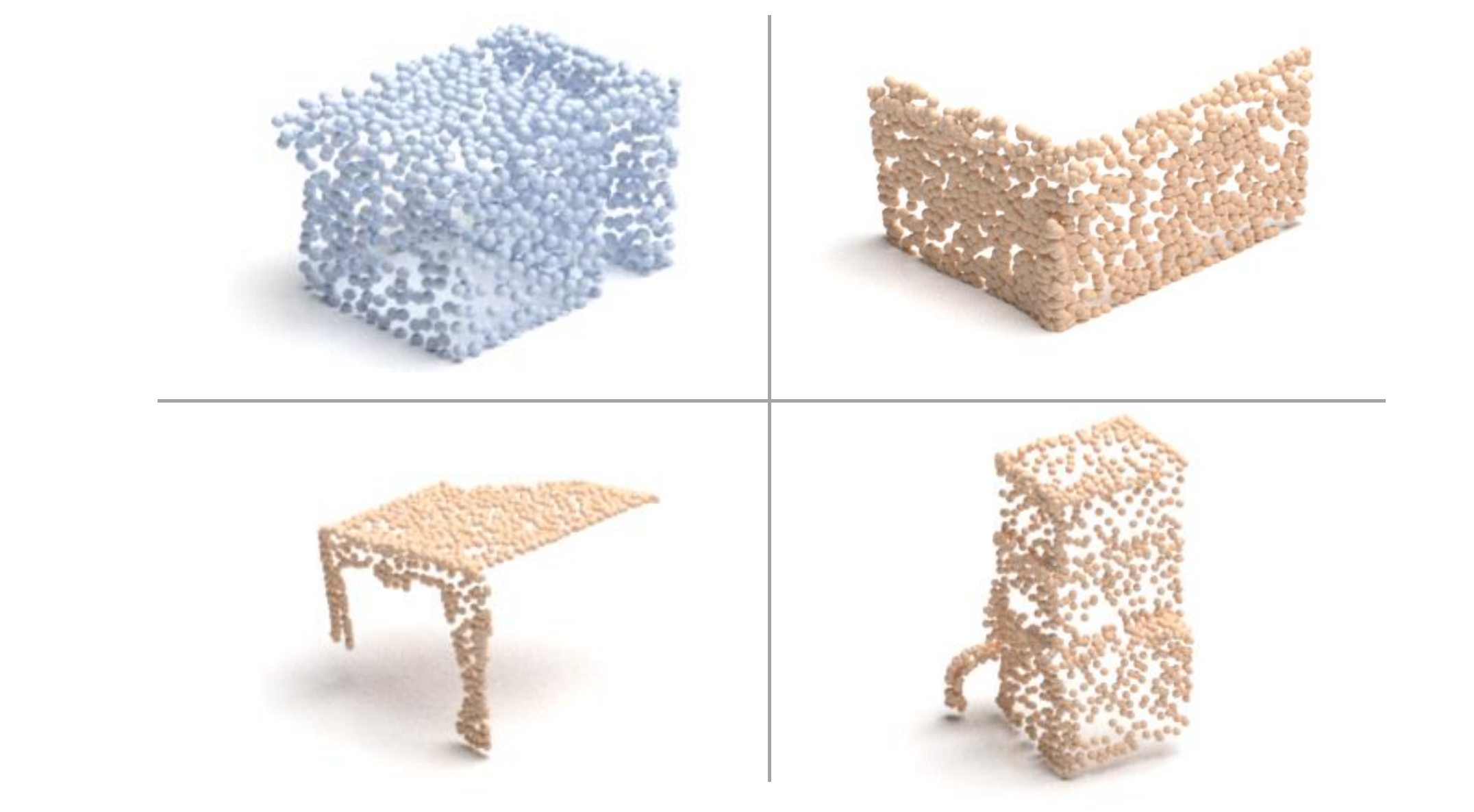}
    \caption{Samples from \textit{cabinet}. First row: intra-class variability between domains (ModelNet and ScanNet). Second row:  intra-class variability in  ScanNet.} 
    \label{fig:intra_variation}
\end{figure}

We first seek to expand the set of easy examples $\mathcal{E}$ by finding in the feature space of $\Phi_{rec}$ the nearest neighbor in $\mathcal{H}$ for each sample of $\mathcal{E}$ and viceversa. To refine \pl{} for samples in $\mathcal{H}$, we adopt a well-know technique employed for surface registration \cite{deng2018ppf,zeng20173dmatch,spezialetti2019learning} and accept only reciprocal nearest neighbor matches, \ie pairs of samples that are mutually the closest one in the feature space: if one sample $h \in \mathcal{H}$ is the nearest neighbour in $\mathcal{H}$ of a sample $e \in \mathcal{E}$ and $e$ is in turn the nearest neighbor of $h$ in $\mathcal{E}$, we move $h$ to the easy split and label it according to the pseudo-label of $e$. At the end of this procedure we obtain a refined set of easy examples $\mathcal{E}^r$.

We then try to refine the \pl{} for the samples left in $\mathcal{H}$ exploiting $\mathcal{E}^r$. 
In particular, we select $K$-nearest neighbors ($K=3$ in our experiments) in the refined set $\mathcal{E}^r$ for each remaining sample $h \in \mathcal{H}$, and assign the new pseudo-label to $h$ by majority voting. When there is no consensus among the $K$ neighbors, we assign the pseudo-label of the closest one. This produces the refined set of hard examples $\mathcal{H}^r$. 
It is important to note that the entire process is applied offline before the \st{} step, as illustrated in \cref{fig:pipeline}, and the absence of hard thresholds in all the refinement steps facilitates the applicability of the proposed method across datasets.

\subsection{Self-training}
\label{self_training}
When performing synthetic-to-real adaptation and vice-versa, the gap among the two distributions could be large and difficult to reduce even in case of perfect supervision. As a matter of example, \cref{fig:intra_variation} shows  how shapes, such as \textit{cabinets}, may look  very  different across domains (first row) as well as  within a domain (second row). A high intra-class variability can be somehow dealt with in a supervised setting, but it is harder to handle when noisy supervision in the form of \pl{} must be used.
Hence, in this setting, it is difficult for a neural network to find common features for shapes belonging to the same class across domains.
We address this issue by adopting domain-specific classification heads together with online \pl{} refinement while performing the \st{} step that concludes our pipeline.

\begin{figure}[t]
    \centering
    \includegraphics[trim={0cm 2.5cm 0cm 1.5cm}, clip, width=0.9\linewidth]{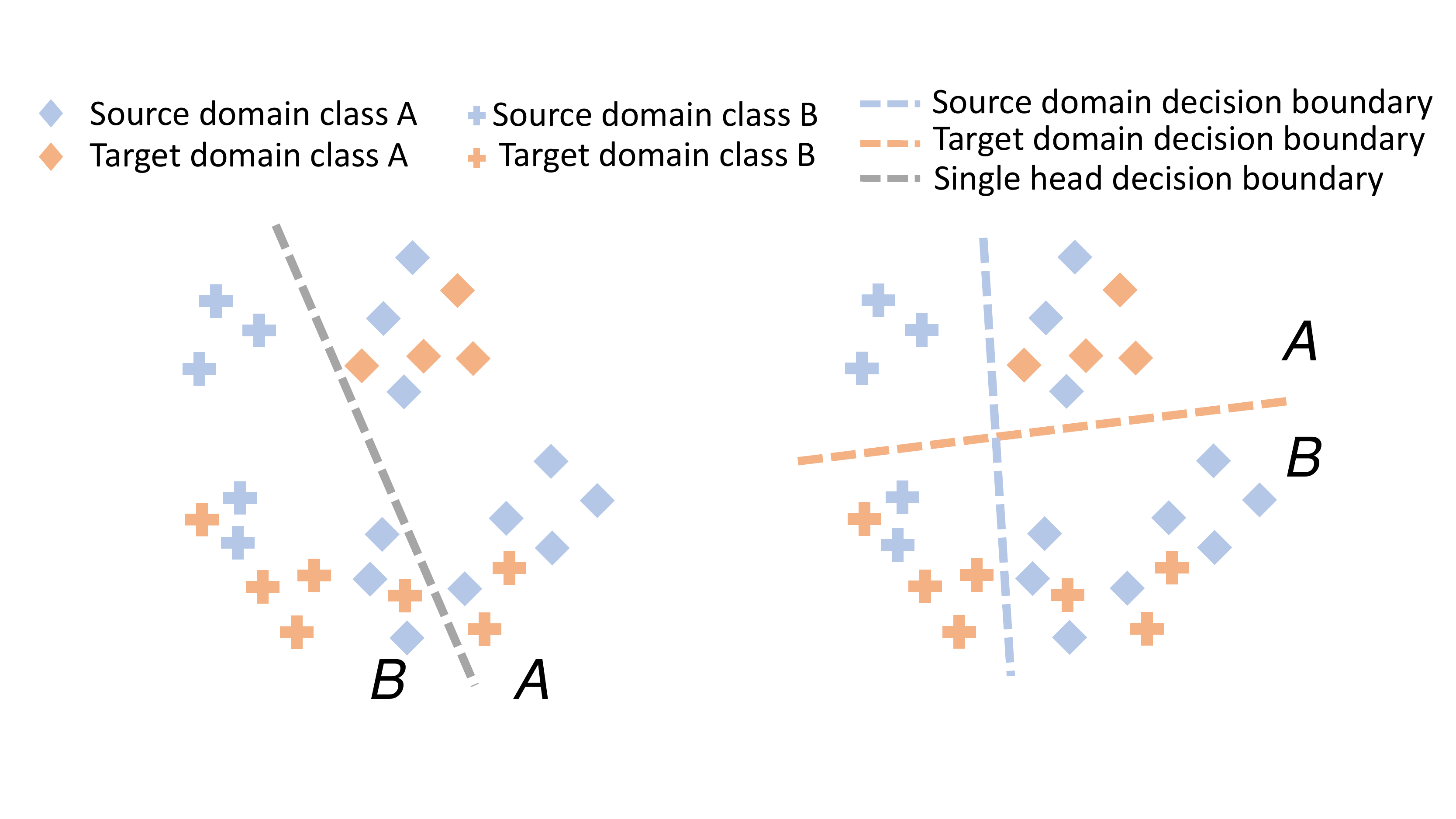}
    \caption{Single head vs domain-specific classification heads. When a single head is deployed (left side), it may not be possible to find a linear decision boundary that correctly classifies both classes for the source and target domain. 
    When domain-specific classification heads are deployed, the model can focus on each domain separately and learn more effective decision boundaries (right side).}
    \label{fig:decision_boundaries}
\end{figure}
\textbf{Domain-specific classification heads.} To tackle intra-class variability across domains, we deploy a shared encoder $\Phi_{cls}$, initialized using the weights from $\Phi_{rec}$, and attach two domain-specific classification heads, $\Psi_s$ and $\Psi_t$, for the source and target domain, respectively.
The benefit of having a target-specific head (right) versus a single head trained on both domains (left) is highlighted in \cref{fig:decision_boundaries}.

When using a single classification head, the model tries to separate classes for both domains simultaneously, which may lead to a non-optimal decision boundary. Indeed,
due to the high intra-class variability across the domains  it is not possible to find a single decision boundary to correctly separate all samples for both, leading to some wrongly classified samples.
When employing two domain-specific heads, instead, the model can learn two more effective boundaries.

Although domain-specific classification heads have been already explored in UDA for image classification \cite{Ren_2021}, in \algoname{} we can apply them in a unique way, which makes them more effective, as it will been shown in our ablation studies. In particular, we train the first head $\Psi_s$ on the source domain mixed with $\mathcal{E}^r$, while we supervise the second head using target data only, \ie{} both $\mathcal{E}^r$ and $\mathcal{H}^r$.
By doing this, we force both heads to correctly classify the easy split, which contains the most confident predictions, \ie the set of target samples already more aligned with the source domain. Thereby, training with this strategy does not reduce performance on the source domain while it enforces a partial feature alignment across domains. At the same time, by only feeding target data to $\Psi_t$, we let the model define target-specific boundaries, alleviating the negative impact of the intra-class variability across domains. 

\textbf{Online \pl{} refinement. }
Even when using domain-specific classification heads, the intra-class variability on the target domain can still affect the model performance. To deal with this issue   we adopt an online \pl{} refinement and weighting strategy.
The key intuition behind online refinement is that, as training progresses, our classifier better learns how to classify the target domain thanks to the pseudo-labels and thus we can progressively improve the \pl{} by exploiting such freshly learned knowledge.
Purposely, we exploit the mean teacher \cite{ema} of our model in order to obtain online \pl{} $\tilde{y}_{t}$: 
\begin{equation}
   		\tilde{y}_{t} =  \Lambda ( \tilde{\Psi}_s(\tilde{\Phi}_{cls}(\mathbf{x}_t)) + \tilde{\Psi}_t(\tilde{\Phi}_{cls}(\mathbf{x}_t)) )
\label{eq:ema_pl}
\end{equation}
It is important to note that $\tilde{\Phi}_{cls}$, $\tilde{\Psi}_s$, and $\tilde{\Psi}_t$ are never updated trough gradient back-propagation, as they consists of simple temporal exponential moving averages (EMA )of their student counterparts \cite{ema}. 
At each training step, we feed one batch of samples from $\{\mathcal{S}, \mathcal{E}^r\}$  and one batch from $\{\mathcal{E}^r, \mathcal{H}^r\}$ to train $\Phi_{cls} \circ \Psi_s$ and $\Phi_{cls} \circ \Psi_t$, respectively. 
As for the source classifier, we train it by the standard cross-entropy loss with labels for the source domain and the \pl{} $\hat{y}_{t}$ obtained from the refinement step for $\mathcal{E}^r$:
\begin{equation}
    \mathcal{L}_s = \mathcal{L}_{ce}(\mathbf{p}_s,\mathbf{y}_s) + \mathcal{L}_{ce}(\mathbf{p}_t, \hat{\mathbf{y}}_{t})
    \label{eq:loss_source}
\end{equation}
We instead exploit both the refined  ($\hat{\mathbf{y}}_{t}$) and the on-line ($\tilde{\mathbf{y}}_{t}$) \pl{} when training the target classifier:
\begin{equation}
    \mathcal{L}_t = (1-\alpha_{it}) z_{t} \mathcal{L}_{ce}(\mathbf{p}_t,\hat{\mathbf{y}}_t) + \alpha_{it} z_{t} \mathcal{L}_{ce}(\mathbf{p}_t, \tilde{\mathbf{y}}_{t})
    \label{eq:loss_target}
\end{equation}
where $\alpha_{it}$ is a  weighting factor that starts from 0 (use only refined \pl{}) and increases at every iteration up to 1 (use only on-line \pl{}). Intuitively, when \st{} starts, we trust the previously refined \pl{} and thus give more weight to the first term of \cref{eq:loss_target} as the mean teacher is not reliable yet. As training goes on, we progressively trust more the output of the mean teacher, \ie $\tilde{y}_{t}$, and so give more weight to the second term.
$z_{t}$ is instead a weighting factor that accounts for the plausibility of the pseudo-label. $z_{t}$ is computed for each target sample exploiting once again the embedding of $\Phi_{rec}$. In particular, before the \st{} step, we compute the class-wise prototypes $\eta^{(k)} \in \mathbb{R}^d$ as the class-wise average of the target features in the easy split:
\begin{equation}
    \eta^{(k)} = \frac{\sum_{x_t\in \mathcal{E}^r} \Phi_{rec}(x_t) * \mathbbm{1}(\hat{y}_{t,k} == 1)}
    {\sum_{x_t\in \mathcal{E}^r} \mathbbm{1}(\hat{y}_{t,k} == 1)}
    \label{eq:prototype_init}
\end{equation}
where $\hat{y}_{t,k}$ is the $k$-entry in $\hat{y}_{t}$.
We only consider $\mathcal{E}^r$ as it contains the most reliable \pl{} for the target domain. We then obtain the confidence score for each sample by simply computing the softmax of the opposite of the distance between the current embedding of $x_t$ and the prototype of the class $k$ assigned to it in its online pseudo-label, $k=\arg \max_{k'} \tilde{y}_{t,k'}$:
\begin{equation}
    z_{t} = \frac{\exp (-\|\tilde{\Phi}_{cls}(\mathbf{x}_t)-\eta^{(k)}\|_2)}
    {\sum_{k'} \exp (-\|\tilde{\Phi}_{cls}(\mathbf{x}_t)-\eta^{(k')}\|_2)}.
    \label{eq:soft_label}
\end{equation}
Hence, $z_{t}$ forces the loss to ignore samples which are far from the expected prototype. 
In fact, when  a sample has a very different representation from the expected class prototype, either the pseduo-label is wrong or the input point cloud is an outlier in the target distribution, and dynamically weighting it less in the \st{} process allows for learning a better classifier.

\begin{table*}[h]
   \centering
   \setlength{\tabcolsep}{4.3mm}
   \scalebox{0.7}{
   \begin{tabular}{p{7.25em}|ccccccc}
      \multicolumn{1}{c}{} &       &       &       &       &       &       &  \\
      \toprule
      \multirow{2}[2]{*}{\textbf{Method}} & \multicolumn{1}{c}{\textbf{ModelNet\ to}} & \multicolumn{1}{c}{\textbf{ModelNet to}} & \multicolumn{1}{c}{\textbf{ShapeNet to}} & \multicolumn{1}{c}{\textbf{ShapeNet to}} & \multicolumn{1}{c}{\textbf{ScanNet to}} & \multicolumn{1}{c}{\textbf{ScanNet to}} & \multicolumn{1}{c}{\multirow{2}[2]{*}{\textbf{Avg}}} \\
      \multicolumn{1}{c|}{} & \multicolumn{1}{c}{\textbf{ShapeNet}} & \multicolumn{1}{c}{\textbf{ScanNet}} & \multicolumn{1}{c}{\textbf{ModelNet}} & \multicolumn{1}{c}{\textbf{ScanNet}} & \multicolumn{1}{c}{\textbf{ModelNet}} & \multicolumn{1}{c}{\textbf{ShapeNet}} &  \\
      \midrule
      No Adaptation & 80.2  & 43.1  & 75.8  & 40.7  & 63.2  & 67.2    & 61.7 \\
      \midrule
      PointDAN \cite{qin2019pointdan} & 80.2  & 45.3  & 71.2  & 46.9  & 59.8  & 66.2  & 61.6 \\
      DefRec \cite{achituve2021self}   & 80.0    & 46.0    & 68.5  & 41.7  & 63.0    & 68.2  & 61.2 \\
      DefRec+PCM \cite{achituve2021self}  & 81.1  & 50.3  & 54.3  & 52.8  & 54.0    & 69.0    & 60.3 \\
      3D Puzzle \cite{alliegro2021joint}  & \textbf{81.6}  & 49.7  & 73.6  & 41.9  & 65.9  & 68.1  & 63.5 \\
        RefRec (Ours)  & 81.4 & \textbf{56.5} & \textbf{85.4}        & \textbf{53.3}        & \textbf{73.0} & \textbf{73.1}      & \textbf{70.5}       \\
        \midrule
        Oracle      & 93.2 & 64.2 & 95 &  64.2 & 95.0 & 93.2 \\
      \bottomrule
     \end{tabular}%
   }
\caption{Shape classification accuracy (\%) on the PointDA-10 dataset. For each method, we report the average results on three runs. Best result on each column is in bold.}  
\label{tab:tab1}%
 \end{table*}%
\begin{table}[t]
   \centering
   \scalebox{0.70}{
    \renewcommand{\tabcolsep}{16pt}
   \begin{tabular}{p{10em}|cccccc}
      \multicolumn{1}{c}{} &       &\\
      \toprule
      \multirow{2}[2]{*}{\textbf{Method}} & \multicolumn{1}{c}{\textbf{ModelNet to}}\\
      \multicolumn{1}{c|}{} & \multicolumn{1}{c}{\textbf{ScanObjectNN}} \\
      \midrule
      No Adaptation & 49.6  & \\
      PointDAN \cite{qin2019pointdan}    & 56.4 \\
      3D Puzzle \cite{alliegro2021joint} & 58.5 \\
      \midrule
        RefRec (Ours)  &  \textbf{61.3}  \\

      \bottomrule
     \end{tabular}%
   }
 \caption{Shape classification accuracy (\%) on the ScanObjectNN dataset. For each method, we report the average results on three runs. Best result is in bold.} 
 \label{tab:scanobj}
 \end{table}%
\section{Experiments}
\label{sec:experiments}
We evaluate our method on two standard datasets for point cloud classification: PointDA-10 \cite{qin2019pointdan} and ScanObjectNN \cite{uy2019revisiting}.

\textbf{PointDA-10. }
PointDA-10 is composed by subsets of three widely adopted datasets for point cloud classification: ShapeNet \cite{chang2015shapenet},  ModelNet40 \cite{modelnet} and ScanNet \cite{dai2017scannet}.   
The subsets share the same ten classes and can be used to define six different pairs of source/target domains, which belongs to three different adaptation scenarios: real-to-synthetic, synthetic-to-synthetic and synthetic-to-real, with the last one arguably the most relevant to practical applications.
ModelNet-10 contains 4183 train samples and 856 tests samples of 3D synthetic CAD models. ShapeNet-10 is a synthetic dataset alike, but exhibits more intra-class variability compared to ModelNet-10. It consists of 17,378 train samples and 2492 test samples. ScanNet-10 is the only real datasets and contains 6110 train and 1769 test samples. ScanNet-10 has been obtained from RGB-D scans of real-world indoor scenes. Due to severe occlusions and noise in the registration process, ScanNet-10 is hard to address even by  standard supervised learning, which renders the associated synthetic-to-real UDA setting very challenging.
Following previous 3D DA work \cite{qin2019pointdan, alliegro2021joint, achituve2021self}, we uniformly sample 1024 points from each 3D shape for training and testing.

\textbf{ScanObjectNN. }
ScanObjectNN is a real-world dataset composed by 2902 3D scans from 15 categories. 
Similarly to ScanNet-10, it represents a challenging scenario due to the high diversity with respect to synthetic datasets and the presence of artifacts such as non-uniform point density, missing parts and occlusions. 
Several variants of the ScanObjectNN dataset are provided. As in \cite{alliegro2021joint}, we select the OBJ\_ONLY version which contains only foreground vertices, and test in the synthetic-to-real setting ModelNet40 to ScanObjectNN using the 11 overlapping classes.

\subsection{Implementation details}
As done in all previous 3D DA methods \cite{qin2019pointdan, alliegro2021joint, achituve2021self}, we use the well-known PointNet \cite{qi2017pointnet} architecture. In particular, we use the standard PointNet  proposed for point cloud classification for all our backbones $\Phi^{w}_{cls}$, $\Phi_{cls}$, and $\Phi_{rec}$. It produces a 1024 dimensional global feature representation for each input point cloud.
We train the reconstruction network for 1000 epochs in the unsupervised \pt{} step \cite{Achlioptas2018LearningRA}, while we train only for 25 epochs when training classification networks in each step of our pipeline.
We set to 0.0001 both learning rate and weight decay. We train with batch size 16 using AdamW \cite{loshchilov2018decoupled} with cosine annealing \cite{SGDR} as optimizer.
The framework is implemented in PyTorch \cite{pytorch}, and is available at {\small{\url{https://github.com/CVLAB-Unibo/RefRec}}}. At test time, we use the target classifier $\Phi_{cls} \circ \Psi_t \circ \Lambda$ in the target domain.

 \begin{table*}[h!]
  \centering
  \setlength{\tabcolsep}{3.3mm}
  \scalebox{0.8}{
  \begin{tabular}{p{5.25em}cccccccc}
      \multicolumn{1}{c}{} &       &       &       &       &       &       &  \\
      \toprule
      \multicolumn{2}{c}{\textbf{Experiment}} & \multicolumn{1}{c}{\textbf{ModelNet\ to}} & \multicolumn{1}{c}{\textbf{ModelNet to}} & \multicolumn{1}{c}{\textbf{ShapeNet to}} & \multicolumn{1}{c}{\textbf{ShapeNet to}} & \multicolumn{1}{c}{\textbf{ScanNet to}} & \multicolumn{1}{c}{\textbf{ScanNet to}} & \multicolumn{1}{c}{\multirow{2}[2]{*}{\textbf{Avg}}} \\
      \multicolumn{1}{c|}{Training data} & \multicolumn{1}{c|}{Offline ref.} & \multicolumn{1}{c}{\textbf{ShapeNet}} & \multicolumn{1}{c}{\textbf{ScanNet}} & \multicolumn{1}{c}{\textbf{ModelNet}} & \multicolumn{1}{c}{\textbf{ScanNet}} & \multicolumn{1}{c}{\textbf{ModelNet}} & \multicolumn{1}{c}{\textbf{ShapeNet}} &  \\
      \midrule
        \multicolumn{1}{c|}{$\mathcal{S}, \mathcal{E}, \mathcal{H}$} & \multicolumn{1}{c|}{}  & \textbf{82.2} & 51.7 & 80.4 & 43.4 & 64.7 & \textbf{70.0} & 65.4       \\
        \multicolumn{1}{c|}{$\mathcal{S}, \mathcal{E}$} & \multicolumn{1}{c|}{}  & 82.8 & 49.6 & 79.0 & 43.8 & 64.1 & 69.8 & 64.9       \\
        \multicolumn{1}{c|}{$\mathcal{S}, \mathcal{E}^r, \mathcal{H}^r$} & \multicolumn{1}{c|}{\checkmark} & 79.1 & \textbf{57.3} & \textbf{85.6} & \textbf{50.7} & \textbf{70.1} & 69.1 & \textbf{68.7}     \\

      \bottomrule
     \end{tabular}%
  }
  \caption{Ablation study on the effect of offline refinement. We report the average shape classification accuracy (\%) on three runs.}
  \label{tab:refine}%
\end{table*}
 
 \begin{table*}[h!]
  \centering
  \setlength{\tabcolsep}{2.5mm}
  \scalebox{0.8}{
  \begin{tabular}{c|cccccccccc}
        \multicolumn{1}{c}{} &       &       &       &       &       &       &  \\
      \toprule
      \multicolumn{4}{c}{\textbf{Experiment}}     & \multicolumn{1}{c}{\textbf{ModelNet\ to}} & \multicolumn{1}{c}{\textbf{ModelNet to}} & \multicolumn{1}{c}{\textbf{ShapeNet to}} & \multicolumn{1}{c}{\textbf{ShapeNet to}} & \multicolumn{1}{c}{\textbf{ScanNet to}} & \multicolumn{1}{c}{\textbf{ScanNet to}} & \multicolumn{1}{c}{\multirow{2}[2]{*}{\textbf{Avg}}} \\
      \multicolumn{1}{c|}{$\Psi_s$} & \multicolumn{1}{c|}{$\Psi_t$} & \multicolumn{1}{c|}{EMA} & \multicolumn{1}{c|}{Online ref.} & \multicolumn{1}{c}{\textbf{ShapeNet}} & \multicolumn{1}{c}{\textbf{ScanNet}} & \multicolumn{1}{c}{\textbf{ModelNet}} & \multicolumn{1}{c}{\textbf{ScanNet}} & \multicolumn{1}{c}{\textbf{ModelNet}} & \multicolumn{1}{c}{\textbf{ShapeNet}} &  \\
      \midrule
        $\mathcal{S}$              & \multicolumn{1}{c|}{$\mathcal{E}^r, \mathcal{H}^r$}   & \multicolumn{1}{c|}{} & \multicolumn{1}{c|}{}                    & 79.5 & 55.3 & 84.7 & 49.3 & 72.0 & 68.5 & 68.2       \\
        $\mathcal{S}, \mathcal{E}^r$ & \multicolumn{1}{c|}{$\mathcal{E}^r, \mathcal{H}^r$} & \multicolumn{1}{c|}{} & \multicolumn{1}{c|}{}                  & 79.3 & 56.9 & 84.7 & 51.6 & 71.7 & 69.0 &68.9       \\
        $\mathcal{S}, \mathcal{E}^r$ & \multicolumn{1}{c|}{$\mathcal{E}^r, \mathcal{H}^r$} & \multicolumn{1}{c|}{\checkmark} & \multicolumn{1}{c|}{} & 80.3 & 54.2 & 83.2 & 52.7 & 72.8 & 71.6 & 69.1     \\
        $\mathcal{S}, \mathcal{E}^r$ & \multicolumn{1}{c|}{$\mathcal{E}^r, \mathcal{H}^r$} & \multicolumn{1}{c|}{\checkmark} & \multicolumn{1}{c|}{\checkmark}        & \textbf{81.4} & \textbf{56.5} & \textbf{85.4} & \textbf{53.3} & \textbf{73.0} & \textbf{73.1} & \textbf{70.5}     \\
    
      \bottomrule                                                   
     \end{tabular}%
  }
\caption{Ablation study on the effect of the self-training strategy and online refinement. We report the average shape classification accuracy (\%) on three runs.}
\label{tab:selftraining}%
\end{table*}%
 
\begin{table*}[h!]
\centering
\setlength{\tabcolsep}{3.3mm}
\scalebox{0.85}{
\begin{tabular}{p{8.25em}|ccccccc}
        \multicolumn{1}{c}{} &       &       &       &       &       &       &  \\
      \toprule
      \multirow{2}[2]{*}{\textbf{Method}} & \multicolumn{1}{c}{\textbf{ModelNet\ to}} & \multicolumn{1}{c}{\textbf{ModelNet to}} & \multicolumn{1}{c}{\textbf{ShapeNet to}} & \multicolumn{1}{c}{\textbf{ShapeNet to}} & \multicolumn{1}{c}{\textbf{ScanNet to}} & \multicolumn{1}{c}{\textbf{ScanNet to}} & \multicolumn{1}{c}{\multirow{2}[2]{*}{\textbf{Avg}}} \\
      \multicolumn{1}{c|}{} & \multicolumn{1}{c}{\textbf{ShapeNet}} & \multicolumn{1}{c}{\textbf{ScanNet}} & \multicolumn{1}{c}{\textbf{ModelNet}} & \multicolumn{1}{c}{\textbf{ScanNet}} & \multicolumn{1}{c}{\textbf{ModelNet}} & \multicolumn{1}{c}{\textbf{ShapeNet}} &  \\
      \midrule
       No Adaptation & 80.2  & 43.1  & 75.8& 40.7  & 63.2  & 67.2    & 61.7 \\
       Warm-up                  & 81.3 & 51.4 & 78.9 & 43.8 & 59.7 & 67.5 & 63.7    \\
       Multi-task       & 80.6 & 45.4 & 78.9 & 46.0 & 63.9 & 67.4 & 63.7    \\
       Self-train multi-task    & 81.2 & 46.9 & 76.3 & 47.7 & 66.0 & 66.5 & 64.1   \\
      RefRec (Ours)             & \textbf{81.4} & \textbf{56.5} & \textbf{85.4} & \textbf{53.3} & \textbf{73.0} & \textbf{73.1} & \textbf{70.5} \\
\bottomrule                                                   
\end{tabular}%
}
\caption{Ablation study on the effect of pre-training. We report the average shape classification accuracy (\%) on three runs.}
\label{tab:warmup}%
\end{table*}%

\subsection{Results}
\label{subsec:results}
We report and discuss here the results of \algoname{}, and compare its performance against previous work as well as the baseline method trained on the source domain and tested on the target domain (referred to as No Adaptation). For each experiment, we provide the mean accuracy obtained on three different seeds. Since in UDA target annotations are not available, we never use target labels to perform model selection and we always select the model that gives the best result on the validation set of the source dataset.

\textbf{PointDA-10. }
We summarize results for each benchmark in \cref{tab:tab1}. 
Overall, our proposal improves by a large margin the previous state-of-the-art methods. Indeed, on average we obtain 70.5\% against the 63.5\% obtained by 3D puzzle \cite{alliegro2021joint}, which is an improvement of 7\% in terms of accuracy. 
From \cref{tab:tab1}, it is also possible to observe how our method is consistently better than previous works in the synthetic-to-real adaptation scenario, which we consider the most important for practical applications.
Compared to DefRec+PCM \cite{achituve2021self}, which obtains 50.3 in the \mtosc{} setting, we improve by 6.2\%. As regards \stosc{}, we obtain 53.3, surpassing by 0.5\% DefRec+PCM again.
Moreover, we highlight how RefRec seems to be the only framework able to generalize well to all adaptation scenarios.
In fact, when comparing our proposal to DefRec+PCM which was the strongest method for the synthetic-to-real case, we also improve by a large margin in cases such as \stom{} and \sctom{}, where DefRec+PCM seems to fail. 
Finally, the ability of RefRec to handle large distributions gaps is also confirmed by the large improvements in the real-to-synthetic cases. Indeed, we observe a +7.1\% improvement for \sctom{} and +4.1\% for \sctos{}. 

\textbf{ScanObjectNN.}
In \cref{tab:scanobj} We report the results for the challenging ModelNet$\xrightarrow{}$ScanObjectNN adapation task. On this challenging benchmark, we achieve 61.3\%, which is 2.8\% better that the previous state-of-the-art result.

\subsection{Ablation studies}
To validate the importance of our design choices, we conduct some ablation studies on both the \pl{} refinement process and the \st{} strategy.

\textbf{Pseudo labels refinement}.
In \cref{tab:refine}, we show the effect of our refinement process. When performing \st{} with the initial, unrefined \pl{} produced by the classifier $\Psi_s$ after the warm-up step (first row), we obtain an overall accuracy of 65.4\%. Conversely,  when we apply our descriptor matching approach aimed at \pl{} refinement (third row), the accuracy increases to 68.7\%. This confirms our intuition that using the reconstruction network allows to capture similarities among shapes in feature space and consequently to improve the  \pl{}. Moreover, we compare \st{} using the most-confident \pl{} only (second row) against \st{} with the refined \pl{} (third row).
The improvement given by the refinement approach (+3.8\%) suggests that only using the most confident \pl{} is not enough to reach good performance.

\textbf{Self-training strategy}. In \cref{tab:selftraining}, we show the effectiveness of our strategy to perform \st{} and ablate our design choices comparing with other reasonable alternatives.
When deploying the domain-specific classification heads, and training $\Psi_s$ solely with source data (first row), results are worse then when we train $\Psi_s$ with both source and $\mathcal{E}^r$ (second row).
This is more evident for the synthetic-to-real adaptation and vice versa, where partial alignment in feature space is more difficult to attain. Indeed in all four cases, forcing $\Psi_s$ to correctly classify the target easy split is beneficial. On the other hand, for the synthetic-to-synthetic case, performances remain stable. This is an expected behaviours since the decision boundaries in these easy adaptation scenarios should not vary significantly across domains.
Finally, in the last two rows, we ablate the effect of the mean teacher and the online refinement, respectively. The mean teacher only gives a marginal contribution (+0.2\%), while its combination with our online refinement accounts for a 1.4\% improvement.

\textbf{Warm-up vs SSL}. Finally, we aim to shed some light on the importance of unsupervised pre-training, \ie{} warm-up, compared to SSL, which is so far the most studied approach to UDA for point cloud classification. In \cref{tab:warmup}, we compare our warm-up step (second row), which exploits unsupervised pre-training, with a multi-task approach as done in \cite{alliegro2021joint, achituve2021self} (multi-task, third row), where the SSL task of shape reconstruction is solved by an auxiliary head. For fair comparison, we adopt in both cases our data augmentation in the synthetic-to-real setting, and train the multi-task architecture for 150 epochs, as done in \cite{alliegro2021joint} and \cite{achituve2021self}, since no pre-training is applied.
Although a simple comparison between such baselines does not establish a clear winner (63.7 on average in both cases), we observe a remarkable difference after the \st{} stage.
Indeed, when comparing the classifier self-trained with \pl{} obtained with the multi-task approach (fourth row) against the classifier self-trained with refined \pl{} (last row), and applying the mean teacher in both cases, we observe a remarkable gap (+7\%).

\section{Conclusion}
\label{sec:conclusion}
In this work, we improved the state of the art in UDA for point cloud classifications. We showed how solving 3D UDA by means of \st{} with supervision from robust \pl{} is a superior paradigm with respect to the established way of tackling it by multi-task learning. Key contributions we make are effective ways to refine \pl{}, offline and online, by leveraging shape descriptors learned to solve shape reconstruction on both domains, as well as a carefully designed \st{} protocol based on domain-specific classification heads and improved supervision by an evolving mean teacher. We hope our results will call for more explorations around the use of \pl{} and \st{} in this emerging area of research. 
In future for future work, we plan to investigate novel methods to perform the warm-up step using metric learning to construct a more discriminative feature space.
\section{Acknowledgment}
The authors would like to thank Injenia Srl for supporting this research.

{\small
\bibliographystyle{ieee_fullname}
\bibliography{egbib}

\begin{thebibliography}{10}\itemsep=-1pt

\bibitem{achituve2021self}
Idan Achituve, Haggai Maron, and Gal Chechik.
\newblock Self-supervised learning for domain adaptation on point clouds.
\newblock In {\em Proceedings of the IEEE/CVF Winter Conference on Applications
  of Computer Vision}, pages 123--133, 2021.

\bibitem{Achlioptas2018LearningRA}
Panos Achlioptas, O. Diamanti, Ioannis Mitliagkas, and L. Guibas.
\newblock Learning representations and generative models for 3d point clouds.
\newblock In {\em ICML}, 2018.

\bibitem{alliegro2021joint}
Antonio Alliegro, Davide Boscaini, and Tatiana Tommasi.
\newblock Joint supervised and self-supervised learning for 3d real world
  challenges.
\newblock In {\em 2020 25th International Conference on Pattern Recognition
  (ICPR)}, pages 6718--6725. IEEE, 2021.

\bibitem{arras2007using}
Kai~O Arras, Oscar~Martinez Mozos, and Wolfram Burgard.
\newblock Using boosted features for the detection of people in 2d range data.
\newblock In {\em Proceedings 2007 IEEE international conference on robotics
  and automation}, pages 3402--3407. IEEE, 2007.

\bibitem{Bousmalis2016}
Konstantinos Bousmalis, George Trigeorgis, Nathan Silberman, Dilip Krishnan,
  and Dumitru Erhan.
\newblock Domain separation networks.
\newblock In {\em Proceedings of the 30th International Conference on Neural
  Information Processing Systems}, NIPS'16, page 343–351, Red Hook, NY, USA,
  2016. Curran Associates Inc.

\bibitem{chang2015shapenet}
Angel~X Chang, Thomas Funkhouser, Leonidas Guibas, Pat Hanrahan, Qixing Huang,
  Zimo Li, Silvio Savarese, Manolis Savva, Shuran Song, Hao Su, et~al.
\newblock Shapenet: An information-rich 3d model repository.
\newblock {\em arXiv preprint arXiv:1512.03012}, 2015.

\bibitem{maxsquare}
Minghao Chen, Hongyang Xue, and Deng Cai.
\newblock Domain adaptation for semantic segmentation with maximum squares
  loss.
\newblock {\em 2019 IEEE/CVF International Conference on Computer Vision
  (ICCV)}, Oct 2019.

\bibitem{chen2019unpaired}
Xuelin Chen, Baoquan Chen, and Niloy~J Mitra.
\newblock Unpaired point cloud completion on real scans using adversarial
  training.
\newblock In {\em International Conference on Learning Representations}, 2019.

\bibitem{Chen2018}
Yuhua Chen, Wen Li, Christos Sakaridis, Dengxin Dai, and Luc Van~Gool.
\newblock Domain adaptive faster r-cnn for object detection in the wild.
\newblock In {\em 2018 IEEE/CVF Conference on Computer Vision and Pattern
  Recognition}, pages 3339--3348, 2018.

\bibitem{dai2017scannet}
Angela Dai, Angel~X Chang, Manolis Savva, Maciej Halber, Thomas Funkhouser, and
  Matthias Nie{\ss}ner.
\newblock Scannet: Richly-annotated 3d reconstructions of indoor scenes.
\newblock In {\em Proceedings of the IEEE conference on computer vision and
  pattern recognition}, pages 5828--5839, 2017.

\bibitem{dai2017bundlefusion}
Angela Dai, Matthias Nie{\ss}ner, Michael Zoll{\"o}fer, Shahram Izadi, and
  Christian Theobalt.
\newblock Bundlefusion: Real-time globally consistent 3d reconstruction using
  on-the-fly surface re-integration.
\newblock {\em ACM Transactions on Graphics 2017 (TOG)}, 2017.

\bibitem{deng2018ppf}
Haowen Deng, Tolga Birdal, and Slobodan Ilic.
\newblock Ppf-foldnet: Unsupervised learning of rotation invariant 3d local
  descriptors.
\newblock In {\em Proceedings of the European Conference on Computer Vision
  (ECCV)}, pages 602--618, 2018.

\bibitem{ganin2015}
Yaroslav Ganin and Victor Lempitsky.
\newblock Unsupervised domain adaptation by backpropagation.
\newblock In {\em Proceedings of the 32nd International Conference on
  International Conference on Machine Learning - Volume 37}, ICML'15, page
  1180–1189. JMLR.org, 2015.

\bibitem{goyal2021revisiting}
Ankit Goyal, Hei Law, Bowei Liu, Alejandro Newell, and Jia Deng.
\newblock Revisiting point cloud shape classification with a simple and
  effective baseline.
\newblock {\em International Conference on Machine Learning}, 2021.

\bibitem{groueix2018}
Thibault Groueix, Matthew Fisher, Vladimir~G. Kim, Bryan Russell, and Mathieu
  Aubry.
\newblock {AtlasNet: A Papier-M\^ach\'e Approach to Learning 3D Surface
  Generation}.
\newblock In {\em Proceedings IEEE Conf. on Computer Vision and Pattern
  Recognition (CVPR)}, 2018.

\bibitem{pmlr-v80-hoffman18a}
Judy Hoffman, Eric Tzeng, Taesung Park, Jun-Yan Zhu, Phillip Isola, Kate
  Saenko, Alexei Efros, and Trevor Darrell.
\newblock {C}y{CADA}: Cycle-consistent adversarial domain adaptation.
\newblock In Jennifer Dy and Andreas Krause, editors, {\em Proceedings of the
  35th International Conference on Machine Learning}, volume~80 of {\em
  Proceedings of Machine Learning Research}, pages 1989--1998. PMLR, 10--15 Jul
  2018.

\bibitem{hua2016scenenn}
Binh-Son Hua, Quang-Hieu Pham, Duc~Thanh Nguyen, Minh-Khoi Tran, Lap-Fai Yu,
  and Sai-Kit Yeung.
\newblock Scenenn: A scene meshes dataset with annotations.
\newblock In {\em 2016 Fourth International Conference on 3D Vision (3DV)},
  pages 92--101. IEEE, 2016.

\bibitem{hua2018pointwise}
Binh-Son Hua, Minh-Khoi Tran, and Sai-Kit Yeung.
\newblock Pointwise convolutional neural networks.
\newblock In {\em Proceedings of the IEEE Conference on Computer Vision and
  Pattern Recognition}, pages 984--993, 2018.

\bibitem{kang2019contrastive}
Guoliang Kang, Lu Jiang, Yi Yang, and Alexander~G Hauptmann.
\newblock Contrastive adaptation network for unsupervised domain adaptation.
\newblock In {\em Proceedings of the IEEE Conference on Computer Vision and
  Pattern Recognition}, pages 4893--4902, 2019.

\bibitem{kazhdan2003rotation}
Michael Kazhdan, Thomas Funkhouser, and Szymon Rusinkiewicz.
\newblock Rotation invariant spherical harmonic representation of 3 d shape
  descriptors.
\newblock In {\em Symposium on geometry processing}, volume~6, pages 156--164,
  2003.

\bibitem{ltir}
Myeongjin Kim and Hyeran Byun.
\newblock Learning texture invariant representation for domain adaptation of
  semantic segmentation.
\newblock {\em 2020 IEEE/CVF Conference on Computer Vision and Pattern
  Recognition (CVPR)}, Jun 2020.

\bibitem{kouw2019review}
Wouter~M Kouw and Marco Loog.
\newblock A review of domain adaptation without target labels.
\newblock {\em IEEE transactions on pattern analysis and machine intelligence},
  43(3):766--785, 2019.

\bibitem{lee2013pseudo}
D. Lee.
\newblock Pseudo-label: The simple and efficient semi-supervised learning
  method for deep neural networks.
\newblock In {\em International Conference on Machine Learning (ICML)
  Workshop}, 2013.

\bibitem{li2018so}
Jiaxin Li, Ben~M Chen, and Gim~Hee Lee.
\newblock So-net: Self-organizing network for point cloud analysis.
\newblock In {\em Proceedings of the IEEE conference on computer vision and
  pattern recognition}, pages 9397--9406, 2018.

\bibitem{Li2017RevisitingBN}
Yanghao Li, Naiyan Wang, Jianping Shi, Jiaying Liu, and Xiaodi Hou.
\newblock Revisiting batch normalization for practical domain adaptation.
\newblock {\em ArXiv}, abs/1603.04779, 2017.

\bibitem{bdl}
Yunsheng Li, Lu Yuan, and Nuno Vasconcelos.
\newblock Bidirectional learning for domain adaptation of semantic
  segmentation.
\newblock {\em 2019 IEEE/CVF Conference on Computer Vision and Pattern
  Recognition (CVPR)}, Jun 2019.

\bibitem{liu2019relation}
Yongcheng Liu, Bin Fan, Shiming Xiang, and Chunhong Pan.
\newblock Relation-shape convolutional neural network for point cloud analysis.
\newblock In {\em Proceedings of the IEEE/CVF Conference on Computer Vision and
  Pattern Recognition}, pages 8895--8904, 2019.

\bibitem{liu2020closer}
Ze Liu, Han Hu, Yue Cao, Zheng Zhang, and Xin Tong.
\newblock A closer look at local aggregation operators in point cloud analysis.
\newblock In {\em European Conference on Computer Vision}, pages 326--342.
  Springer, 2020.

\bibitem{NIPS2016_ac627ab1}
Mingsheng Long, Han Zhu, Jianmin Wang, and Michael~I Jordan.
\newblock Unsupervised domain adaptation with residual transfer networks.
\newblock In D. Lee, M. Sugiyama, U. Luxburg, I. Guyon, and R. Garnett,
  editors, {\em Advances in Neural Information Processing Systems}, volume~29.
  Curran Associates, Inc., 2016.

\bibitem{Long2017}
Mingsheng Long, Han Zhu, Jianmin Wang, and Michael~I. Jordan.
\newblock Deep transfer learning with joint adaptation networks.
\newblock In {\em Proceedings of the 34th International Conference on Machine
  Learning - Volume 70}, ICML'17, page 2208–2217. JMLR.org, 2017.

\bibitem{SGDR}
Ilya Loshchilov and Frank Hutter.
\newblock {SGDR:} stochastic gradient descent with warm restarts.
\newblock In {\em 5th International Conference on Learning Representations,
  {ICLR} 2017, Toulon, France, April 24-26, 2017, Conference Track
  Proceedings}. OpenReview.net, 2017.

\bibitem{loshchilov2018decoupled}
Ilya Loshchilov and Frank Hutter.
\newblock Decoupled weight decay regularization.
\newblock In {\em International Conference on Learning Representations}, 2019.

\bibitem{pang2021tearingnet}
Jiahao Pang, Duanshun Li, and Dong Tian.
\newblock Tearingnet: Point cloud autoencoder to learn topology-friendly
  representations.
\newblock In {\em Proceedings of the IEEE/CVF Conference on Computer Vision and
  Pattern Recognition}, pages 7453--7462, 2021.

\bibitem{pytorch}
Adam Paszke, Sam Gross, Francisco Massa, Adam Lerer, James Bradbury, Gregory
  Chanan, Trevor Killeen, Zeming Lin, Natalia Gimelshein, Luca Antiga, Alban
  Desmaison, Andreas Kopf, Edward Yang, Zachary DeVito, Martin Raison, Alykhan
  Tejani, Sasank Chilamkurthy, Benoit Steiner, Lu Fang, Junjie Bai, and Soumith
  Chintala.
\newblock Pytorch: An imperative style, high-performance deep learning library.
\newblock In H. Wallach, H. Larochelle, A. Beygelzimer, F. d\textquotesingle
  Alch\'{e}-Buc, E. Fox, and R. Garnett, editors, {\em Advances in Neural
  Information Processing Systems 32}, pages 8024--8035. Curran Associates,
  Inc., 2019.

\bibitem{Pei2018MultiAdversarialDA}
Zhongyi Pei, Zhangjie Cao, Mingsheng Long, and Jianmin Wang.
\newblock Multi-adversarial domain adaptation.
\newblock In {\em AAAI}, 2018.

\bibitem{Pinheiro2018UnsupervisedDA}
Pedro H.~O. Pinheiro.
\newblock Unsupervised domain adaptation with similarity learning.
\newblock {\em 2018 IEEE/CVF Conference on Computer Vision and Pattern
  Recognition}, pages 8004--8013, 2018.

\bibitem{qi2017pointnet}
Charles~R Qi, Hao Su, Kaichun Mo, and Leonidas~J Guibas.
\newblock Pointnet: Deep learning on point sets for 3d classification and
  segmentation.
\newblock In {\em Proceedings of the IEEE conference on computer vision and
  pattern recognition}, pages 652--660, 2017.

\bibitem{qi2017pointnet++}
Charles~Ruizhongtai Qi, Li Yi, Hao Su, and Leonidas~J Guibas.
\newblock Pointnet++: Deep hierarchical feature learning on point sets in a
  metric space.
\newblock In {\em NIPS}, 2017.

\bibitem{qin2019pointdan}
Can Qin, Haoxuan You, Lichen Wang, C.-C.~Jay Kuo, and Yun Fu.
\newblock Pointdan: A multi-scale 3d domain adaption network for point cloud
  representation.
\newblock In H. Wallach, H. Larochelle, A. Beygelzimer, F. d\textquotesingle
  Alch\'{e}-Buc, E. Fox, and R. Garnett, editors, {\em Advances in Neural
  Information Processing Systems 32}, pages 7190--7201. Curran Associates,
  Inc., 2019.

\bibitem{Ren_2021}
Chuan-Xian Ren, Pengfei Ge, Peiyi Yang, and Shuicheng Yan.
\newblock Learning target-domain-specific classifier for partial domain
  adaptation.
\newblock {\em IEEE Transactions on Neural Networks and Learning Systems},
  32(5):1989–2001, May 2021.

\bibitem{rubner2000earth}
Yossi Rubner, Carlo Tomasi, and Leonidas~J Guibas.
\newblock The earth mover's distance as a metric for image retrieval.
\newblock {\em International journal of computer vision}, 40(2):99--121, 2000.

\bibitem{Saito2019}
Kuniaki Saito, Yoshitaka Ushiku, Tatsuya Harada, and Kate Saenko.
\newblock Strong-weak distribution alignment for adaptive object detection.
\newblock In {\em 2019 IEEE/CVF Conference on Computer Vision and Pattern
  Recognition (CVPR)}, pages 6949--6958, 2019.

\bibitem{Saito2018MaximumCD}
Kuniaki Saito, Kohei Watanabe, Y. Ushiku, and T. Harada.
\newblock Maximum classifier discrepancy for unsupervised domain adaptation.
\newblock {\em 2018 IEEE/CVF Conference on Computer Vision and Pattern
  Recognition}, pages 3723--3732, 2018.

\bibitem{salti2010use}
Samuele Salti, Federico Tombari, and Luigi Di~Stefano.
\newblock On the use of implicit shape models for recognition of object
  categories in 3d data.
\newblock In {\em Asian Conference on Computer Vision}, pages 653--666.
  Springer, 2010.

\bibitem{spezialetti2019learning}
Riccardo Spezialetti, Samuele Salti, and Luigi~Di Stefano.
\newblock Learning an effective equivariant 3d descriptor without supervision.
\newblock In {\em Proceedings of the IEEE/CVF International Conference on
  Computer Vision}, pages 6401--6410, 2019.

\bibitem{spezialetti2020learning}
Riccardo Spezialetti, Federico Stella, Marlon Marcon, Luciano Silva, Samuele
  Salti, and Luigi Di~Stefano.
\newblock Learning to orient surfaces by self-supervised spherical cnns.
\newblock {\em Advances in Neural Information Processing Systems}, 33, 2020.

\bibitem{Sun2016DeepCC}
Baochen Sun and Kate Saenko.
\newblock Deep coral: Correlation alignment for deep domain adaptation.
\newblock In {\em ECCV Workshops}, 2016.

\bibitem{ema}
Antti Tarvainen and Harri Valpola.
\newblock Mean teachers are better role models: Weight-averaged consistency
  targets improve semi-supervised deep learning results.
\newblock In {\em Proceedings of the 31st International Conference on Neural
  Information Processing Systems}, NIPS'17, page 1195–1204, Red Hook, NY,
  USA, 2017. Curran Associates Inc.

\bibitem{thomas2019kpconv}
Hugues Thomas, Charles~R Qi, Jean-Emmanuel Deschaud, Beatriz Marcotegui,
  Fran{\c{c}}ois Goulette, and Leonidas~J Guibas.
\newblock Kpconv: Flexible and deformable convolution for point clouds.
\newblock In {\em Proceedings of the IEEE/CVF International Conference on
  Computer Vision}, pages 6411--6420, 2019.

\bibitem{shift}
Antonio Torralba and Alexei~A. Efros.
\newblock Unbiased look at dataset bias.
\newblock In {\em CVPR 2011}, pages 1521--1528, 2011.

\bibitem{adaptsegnet}
Yi-Hsuan Tsai, Wei-Chih Hung, Samuel Schulter, Kihyuk Sohn, Ming-Hsuan Yang,
  and Manmohan Chandraker.
\newblock Learning to adapt structured output space for semantic segmentation.
\newblock {\em 2018 IEEE/CVF Conference on Computer Vision and Pattern
  Recognition}, Jun 2018.

\bibitem{Tzeng2017}
Eric Tzeng, Judy Hoffman, Kate Saenko, and Trevor Darrell.
\newblock Adversarial discriminative domain adaptation.
\newblock In {\em 2017 IEEE Conference on Computer Vision and Pattern
  Recognition (CVPR)}, pages 2962--2971, 2017.

\bibitem{Tzeng2014DeepDC}
Eric Tzeng, Judy Hoffman, Ning Zhang, Kate Saenko, and Trevor Darrell.
\newblock Deep domain confusion: Maximizing for domain invariance.
\newblock {\em ArXiv}, abs/1412.3474, 2014.

\bibitem{uy2019revisiting}
Mikaela~Angelina Uy, Quang-Hieu Pham, Binh-Son Hua, Thanh Nguyen, and Sai-Kit
  Yeung.
\newblock Revisiting point cloud classification: A new benchmark dataset and
  classification model on real-world data.
\newblock In {\em Proceedings of the IEEE/CVF International Conference on
  Computer Vision}, pages 1588--1597, 2019.

\bibitem{Wang_2018}
Mei Wang and Weihong Deng.
\newblock Deep visual domain adaptation: A survey.
\newblock {\em Neurocomputing}, 312:135–153, Oct 2018.

\bibitem{Wang_2019_CVPR}
Tao Wang, Xiaopeng Zhang, Li Yuan, and Jiashi Feng.
\newblock Few-shot adaptive faster r-cnn.
\newblock In {\em Proceedings of the IEEE/CVF Conference on Computer Vision and
  Pattern Recognition (CVPR)}, June 2019.

\bibitem{Wang_2019}
Xudong Wang, Zhaowei Cai, Dashan Gao, and Nuno Vasconcelos.
\newblock Towards universal object detection by domain attention.
\newblock In {\em Proceedings of the IEEE/CVF Conference on Computer Vision and
  Pattern Recognition (CVPR)}, June 2019.

\bibitem{wang2019dynamic}
Yue Wang, Yongbin Sun, Ziwei Liu, Sanjay~E Sarma, Michael~M Bronstein, and
  Justin~M Solomon.
\newblock Dynamic graph cnn for learning on point clouds.
\newblock {\em Acm Transactions On Graphics (tog)}, 38(5):1--12, 2019.

\bibitem{DCAN}
Zuxuan Wu, Xintong Han, Yen-Liang Lin, Mustafa~G{\"o}khan Uzunbas, Tom
  Goldstein, Ser~Nam Lim, and Larry~S. Davis.
\newblock Dcan: Dual channel-wise alignment networks for unsupervised scene
  adaptation.
\newblock In Vittorio Ferrari, Martial Hebert, Cristian Sminchisescu, and Yair
  Weiss, editors, {\em Computer Vision -- ECCV 2018}, pages 535--552, Cham,
  2018. Springer International Publishing.

\bibitem{wu20153d}
Zhirong Wu, Shuran Song, Aditya Khosla, Fisher Yu, Linguang Zhang, Xiaoou Tang,
  and Jianxiong Xiao.
\newblock 3d shapenets: A deep representation for volumetric shapes.
\newblock In {\em Proceedings of the IEEE conference on computer vision and
  pattern recognition}, pages 1912--1920, 2015.

\bibitem{modelnet}
Zhirong Wu, Shuran Song, Aditya Khosla, Fisher Yu, Linguang Zhang, Xiaoou Tang,
  and Jianxiong Xiao.
\newblock 3d shapenets: A deep representation for volumetric shapes.
\newblock In {\em 2015 IEEE Conference on Computer Vision and Pattern
  Recognition (CVPR)}, pages 1912--1920, 2015.

\bibitem{xu2020grid}
Qiangeng Xu, Xudong Sun, Cho-Ying Wu, Panqu Wang, and Ulrich Neumann.
\newblock Grid-gcn for fast and scalable point cloud learning.
\newblock In {\em Proceedings of the IEEE/CVF Conference on Computer Vision and
  Pattern Recognition}, pages 5661--5670, 2020.

\bibitem{yan2020pointasnl}
Xu Yan, Chaoda Zheng, Zhen Li, Sheng Wang, and Shuguang Cui.
\newblock Pointasnl: Robust point clouds processing using nonlocal neural
  networks with adaptive sampling.
\newblock In {\em Proceedings of the IEEE/CVF Conference on Computer Vision and
  Pattern Recognition}, pages 5589--5598, 2020.

\bibitem{yang2018foldingnet}
Yaoqing Yang, Chen Feng, Yiru Shen, and Dong Tian.
\newblock Foldingnet: Point cloud auto-encoder via deep grid deformation.
\newblock In {\em Proceedings of the IEEE Conference on Computer Vision and
  Pattern Recognition}, pages 206--215, 2018.

\bibitem{zeng20173dmatch}
Andy Zeng, Shuran Song, Matthias Nie{\ss}ner, Matthew Fisher, Jianxiong Xiao,
  and Thomas Funkhouser.
\newblock 3dmatch: Learning local geometric descriptors from rgb-d
  reconstructions.
\newblock In {\em Proceedings of the IEEE conference on computer vision and
  pattern recognition}, pages 1802--1811, 2017.

\bibitem{zou2018unsupervised}
Yang Zou, Zhiding Yu, BVK~Vijaya Kumar, and Jinsong Wang.
\newblock Unsupervised domain adaptation for semantic segmentation via
  class-balanced self-training.
\newblock In {\em Proceedings of the European Conference on Computer Vision
  (ECCV)}, pages 289--305, 2018.

\end{thebibliography}
}
\newpage\phantom{Supplementary}
\multido{\i=1+1}{4}{


\section*{Supplementary material}

# RefRec: Pseudo-labels Refinement via Shape Reconstruction for Unsupervised 3D Domain Adaptation
# (*Supplementary Materials*)

Adriano Cardace    Riccardo Spezialetti    Pierluigi Zama Ramirez
Samuele Salti    Luigi Di Stefano
Department of Computer Science and Engineering (DISI)
University of Bologna, Italy
{adriano.cardace2, riccardo.spezialetti, pierluigi.zama}@unibo.it

## 1. Additional Experiments

### 1.1. Additional settings

We report in Tab. 1 additional results for the setting ScanObjectNN→ModelNet and ModelNet→ScanObjectNN. We observe how in both cases the warm-up considerably improve over the baseline, and that our refinement approach yields the larger boost for both settings.

| Method | ModelNet to ScanObjectNN | ScanObjectNN to ModelNet |
|---|---|---|
| No Adaptation | 49.6 | 47.6 |
| Warm up | 54.1 | 58.5 |
| RefRec (Ours) | **61.3** | **68.9** |
| Oracle | 77.9 | 92.2 |

Table 1. Shape classification accuracy (%) on the ScanObjectNN and ModelNet datasets. For each method, we report the average results on three runs. Best result is in bold.

### 1.2. Fine-tuning strategies

As explained in Sec. 3.1, during the warm-up step, we initialize the classifier backbone $\Phi_{cls}^w$ with weights from $\Phi_{rec}$. Then, we attach a new classification head $\Psi_s$ and train the full model. An alternative solution, is to freeze the backbone, using thereby the representation learned by an auto-encoder and fine-tuning only $\Psi_s$. We compare the two approaches in Tab. 2. We first observe that the strategy proposed in Sec. 3.1 yields on average the best result. Interestingly, on the other hand, just fine-tuning the classification head produces the worst result on average. This suggests that the reconstruction auto-encoder embeddings cannot be used directly to classify the target point clouds in a UDA setting and that combining both networks allows to obtain a stronger target classifier.

### 1.3. Refinement examples

We provide here some qualitative examples of refined pseduo-labels for both the ModelNet→ScanNet and the ShapeNet→ScanNet settings in Fig. 1 and Fig. 2 respectively. Before the refinement process explained in Sec 3.2, samples in the hard split $\mathcal{H}$ (first column) have a wrong pseudo-label assigned by the warm-up source classifier $\Omega^w = \Phi_{cls}^w \circ \Psi^w$ (second column). We compute for each sample its nearest neighbour in the easy split $\mathcal{E}$ exploiting the embedding representation of $\Phi_{rec}$ (third column), and update the previously wrong prediction with the pseudo-label of the nearest neighbour (last column).

### 1.4. Failure cases

Finally, we show some failure cases of our refinement approach, and discuss some limitations that we plan to address in future work. First of all, we report failures due to the use of the reconstruction embedding to match shapes. Indeed, as can be seen from Fig. 3, some shapes previously correctly classified by the classifier are associated to misclassified samples. For example, in the first row of Fig. 3, a sample initially classified as *Lamp*, has as its nearest neighbour a shape with class *Plant*. This leads to a wrong update, that can occur every time the closest embeddings in $\mathcal{E}$ belongs to a different class. An alternative approach could be to train the network with metric learning losses, which can be used to learn more effective feature spaces for matching purposes. Another limitation of our approach, is the choice of a single prototype for each class. Indeed, in some adaptation settings such as ScanNet→ShapeNet (with the latter being the dataset with the highest intra-class variability) there is a drop in performance for classes such as *plant* and *lamp*, i.e. classes that exhibits a large variability in appearance. Since prototypes are computed only on the target domain, such variability can lead our loss (Eq. 3 of the main

| Method | ModelNet to ShapeNet | ModelNet to ScanNet | ShapeNet to ModelNet | ShapeNet to ScanNet | ScanNet to ModelNet | ScanNet to ShapeNet | Avg |
|---|---|---|---|---|---|---|---|
| No Adaptation | 80.2 | 43.1 | 75.8 | 40.7 | 63.2 | 67.2 | 61.7 |
| Fine-tune $\Psi_s$ | 81.5 | 47.8 | 76.2 | 46.8 | 49.1 | 58.3 | 60.0 |
| Fine-tune $\Phi_{cls}^w$ and $\Psi_s$ | 81.3 | 51.4 | 78.9 | 43.8 | 59.7 | 67.5 | 63.7 |

Table 2. Full fine-tuning vs Classification head fine-tuning

paper) to ignore some samples that are far from the corresponding class prototype. This issue could be addressed by using more than one prototype for each class.

| Input Shape in $\mathcal{H}$ | Classifier Pseudo-label | Nearest Neighbour in $\mathcal{E}$ | Refined Pseudo-label |
|---|---|---|---|
| 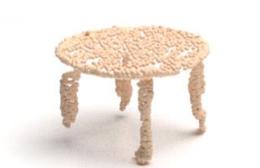 | *Cabinet* | 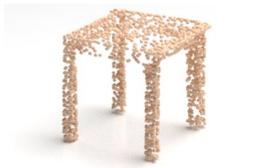 | *Table* |
| 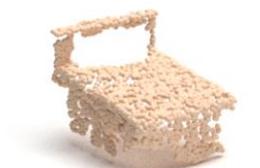 | *Chair* | 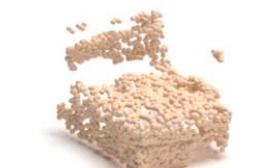 | *Bed* |
| 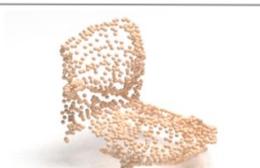 | *Bed* | 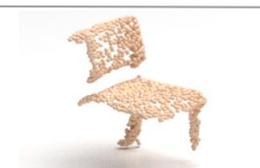 | *Chair* |
| 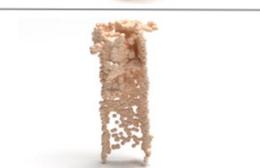 | *Lamp* | 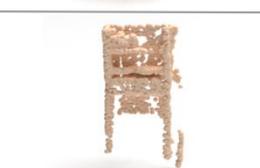 | *Bookshelf* |

Figure 1. Matching examples for ModelNet→ScanNet. Green and red colors correspond to the GT and the prediction respectively.

| Input Shape in $\mathcal{H}$ | Classifier Pseudo-label | Nearest Neighbour in $\mathcal{E}$ | Refined Pseudo-label |
|---|---|---|---|
| 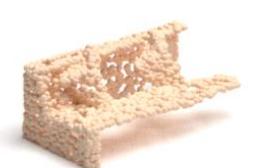 | *Table* | 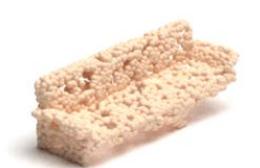 | *Sofa* |
| 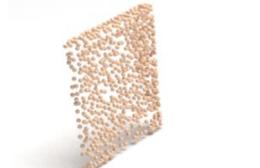 | *Bookshelf* | 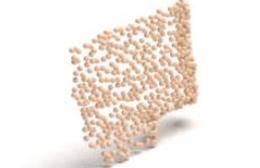 | *Monitor* |
| 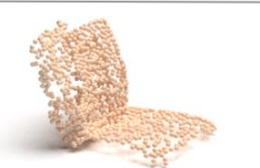 | *Monitor* | 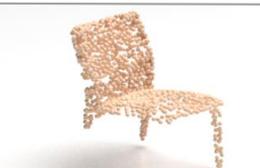 | *Chair* |
| 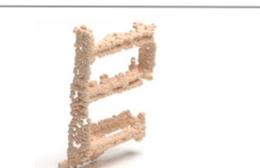 | *Cabinet* | 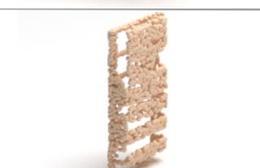 | *Bookshelf* |

Figure 2. Matching examples for ShapeNet→ScanNet. Green and red colors correspond to the GT and the prediction respectively.

| Input Shape in $\mathcal{H}$ | Classifier Pseudo-label | Nearest Neighbour in $\mathcal{E}$ | Refined Pseudo-label |
|---|---|---|---|
| 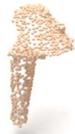 | *Lamp* | 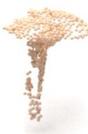 | *Plant* |
| 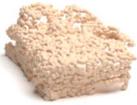 | *Bed* | 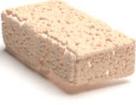 | *Table* |
| 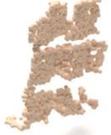 | *Bookshelf* | 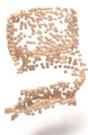 | *Monitor* |
| 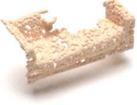 | *Sofa* | 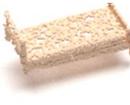 | *Bed* |

Figure 3. Failure cases. When the closest embeddings belongs to a different class, pseudo-labels are wrongly updated. Green and red colors correspond to the GT and the prediction respectively.


}

\end{document}